\documentclass{article}

    \PassOptionsToPackage{numbers, compress}{natbib}

\usepackage[preprint]{neurips_2020}

\usepackage[utf8]{inputenc} 
\usepackage[T1]{fontenc}    
\usepackage{hyperref}       
\usepackage{url}            
\usepackage{booktabs}       
\usepackage{amsfonts}       
\usepackage{nicefrac}       
\usepackage{microtype}      
\usepackage{setspace, longtable, graphicx, hyphenat, hyperref, fancyhdr, ifthen, enumitem, amsmath, setspace}
\usepackage{float}

\title{ChatFDA: Medical Records Risk Assessment}

\author{
  Minh Tran \\
  Robotics Institute \\
  Carnegie Mellon University\\
  \texttt{minht@cs.cmu.edu} \\
  \And
  Charlie Sun \\
  Language Technologies Institute \\
  Carnegie Mellon University \\
  \texttt{cssun@andrew.cmu.edu} \\
  }

\begin{document}
\maketitle
\begin{abstract}
In healthcare, the emphasis on patient safety and the minimization of medical errors cannot be overstated. Despite concerted efforts, many healthcare systems, especially in low-resource regions, still grapple with preventing these errors effectively. This study explores a pioneering application aimed at addressing this challenge by assisting caregivers in gauging potential risks derived from medical notes. The application leverages data from openFDA, delivering real-time, actionable insights regarding prescriptions. Preliminary analyses conducted on the MIMIC-III \cite{mimic} dataset affirm a proof of concept highlighting a reduction in medical errors and an amplification in patient safety. This tool holds promise for drastically enhancing healthcare outcomes in settings with limited resources. To bolster reproducibility and foster further research, the entire codebase underpinning our methodology is accessible on \href{https://github.com/CarolusSolis/2023.hackAuton/tree/main/prescription_checker}{Github}. 
\end{abstract}

\section{Introduction}
Every year in the United States alone, there are 250,000 deaths due to medical errors. These can stem from diagnostic errors, surgeries, patient care infections, and largely (44\%) \cite{adverse} from medications. This situation persists even in a country with a well-developed healthcare system boasting advanced medical facilities, meticulously organized and digitalized records. In low-resource regions, especially in rural areas of developing countries, the challenge of ensuring patient safety intensifies. There are regions where a single nurse may be responsible for an entire village, and they might not have access to standardized medical records. Sometimes, the records are merely handwritten notes or even recollections from the patients themselves.

Knowing that eliminating medical records could save millions live per year, we are motivated to create a tool to leverage the power of LLMs to assure the safety for patients, especially for areas where ressources are limited. We built an application that takes multimodalities medical records input from text, voice, to image then extract/correct any mistake made by human. Based on the medical history and prescription, we also provided risk assessment for each case.

\section{Related Works}

\subsection{Medical Records Correction}
Medical record correction has been a focal point in healthcare informatics for several years. Various studies have delved into the use of Electronic Health Records (EHRs) to bolster the accuracy and reliability of medical data. In the past, probabilistic models have been employed to identify and rectify inconsistencies in medical records, especially in the realm of medication prescriptions. Recently, large language models (like GPT-4 and PaLM) \cite{gpt4, palm} have found applications in general error correction. However, benchmarks specific to medical data remain nonexistent. These studies underscore the paramount importance of precise medical record-keeping, an area our research aspires to further by assessing risks based on these records.

\subsection{Language Models for Medical Data}
The infusion of language models into the medical domain is an emerging trend, further propelled by the inception of advanced models such as Med-PaLM 2 and Med-BERT \cite{med-palm, med-bert}. Such models have been harnessed for diverse tasks including medical text summarization, diagnosis prediction, and drug interaction identification. Yet, there's a noticeable research gap on the efficacy of leveraging LLMs to bolster patient safety, especially in multilingual contexts with limited information. Our study seeks to fill this void by incorporating language models to interpret and authenticate medical notes, especially in environments where expert intervention is scarce. This section offers a snapshot of potential research trajectories and applications of LLMs in medical contexts.

Both of these domains—medical records correction and language models for medical data—shed light on invaluable insights and foundational principles that shape and bolster the ambitions of our research. By marrying elements from these two realms, we endeavor to design an application with the potential to markedly enhance patient safety and curtail medical errors in resource-constrained settings.



\section{Proposed Approach}
\subsection{Pipeline Design}
Our application's architecture is structured as a pipeline comprising several interlinked modules:

\begin{enumerate}
    \item \textbf{Data Collection Module}: This initial module gathers and processes various forms of medical data. While our inputs can range from text to voice and images, we've specifically honed in on medical notes for this experiment due to their reliability as a primary information source for caregivers.
    
    \item \textbf{Data Standardization Module}: Using the MIMIC-III dataset as a foundation, this module standardizes the amassed data, ensuring its consistency and compatibility for further analysis.
    
    \item \textbf{Medical Records Processing Module with GPT-4}: From the standardized data, this module employs the GPT-4 model to segregate the medical records into two distinct sections: prescriptions and medical history. This vital step not only mitigates potential human errors during note-taking but also precisely identifies each patient's prescribed treatments.
    
    \item \textbf{Prescription Analysis Module with openFDA}: The prescription data is relayed to openFDA, which subsequently offers actionable insights. This encompasses potential medication interactions and pertinent treatment guidelines.
    
    \item \textbf{Risk Evaluation and Record Storage Module}: This concluding block synthesizes the insights derived from openFDA with the patient's medical history to generate a comprehensive risk evaluation. Concurrently, it ensures the updated medical records are safely stored in the database, facilitating future retrievals and analyses.
\end{enumerate}

\begin{figure}[H]
    \centering
    \includegraphics[width=15cm]{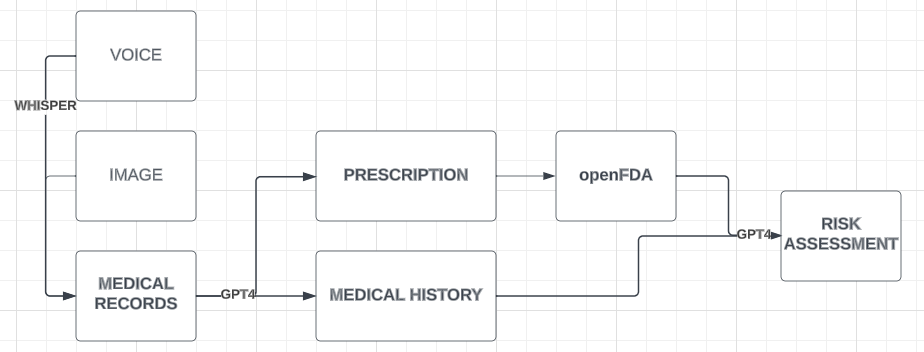}
    \caption{Pipeline Diagram}
    \label{fig:p1}
\end{figure}

\subsection{Prompt Design}

Designing a useful prompt for a language model is critical, especially in high-stakes situations like medical evaluations. In this study, we introduce two separate prompts: 

\begin{enumerate}
    \item The first is to extract relevant medical information from a doctor's notes. This information pertains to a patient's pre-existing conditions, symptoms, and prescribed medications.
    
    \item The second prompt aims to assess the potential risks related to the extracted medical information. In this section, we elaborate on our approach to designing this risk-assessment prompt.
\end{enumerate}

\textbf{Prompt Structure:} The risk assessment prompt comprises three primary sections:

\begin{itemize}
    \item \textbf{Parsed Notes:} This section integrates the parsed information from the doctor's notes.
    
    \item \textbf{Drug Information:} Here, we incorporate drug interactions and warnings retrieved from the FDA database.
    
    \item \textbf{Answer Section:} The language model is tasked to analyze the prescribed treatment, identifying potential drug interactions and assessing potential patient reactions based on their pre-existing conditions. The culmination of this analysis is the evaluation of the treatment's dangerousness on a scale: LOW, MEDIUM, or HIGH. An answer template is also provided to ensure consistency in responses.
\end{itemize}

\textbf{Actual Prompt:}
\begin{verbatim}
"I am a doctor, and I need you to evaluate my prescription:
{parsed_notes}

Drug contexts:
{drug_info_string}

Please answer the following in a concise point format, considering the provided drug context:
- Possible interactions between the prescribed drugs?
- Specific adverse effects of the drugs that relate to the patient's pre-existing conditions
and symptoms?

Conclude your response by assessing the treatment's dangerousness based on interactions and 
adverse effects specific to the patient. Categorize dangerousness as: LOW, MEDIUM, HIGH. 

Your answer should adhere to this format:
* INTERACTIONS:
- <interaction 1>
- <interaction 2>
- ...

* ADVERSE EFFECTS:
- <adverse effect 1>
- <adverse effect 2>
- ...

* DANGEROUSNESS: <LOW / MEDIUM / HIGH>

Include only necessary interactions or adverse effects in your response."
\end{verbatim}

\subsection{User Interface}
The user interface of the application is intentionally designed to be intuitive and user-friendly. It features a series of prompts that guide healthcare workers through the verification process. Initially, users are prompted to select their preferred method of input—image, voice, or text. Based on this selection, the application provides an interface for capturing the image, recording the voice, or typing the text. Once the data is processed and analyzed, a summary report is displayed, and users are prompted to confirm its accuracy. They are also prompted to review the actionable insights generated from openFDA data, allowing for more informed decision-making.

By synergizing an efficient pipeline with user-friendly prompts, our proposed approach aims to offer a robust and accessible application capable of significantly reducing medical errors and improving patient safety, particularly in low-resource settings.



\section{Results Analsysis}


We tested our application on a public sample of the MIMIC-III dataset that contains de-identified health data associated with over 40,000 patients who stayed in critical care units of the Beth Israel Deaconess Medical Center. For this scope of this project, we only test on a small public sample, and focused on medical notes only.

\begin{figure}[H]
    \centering
    \includegraphics[width=12cm]{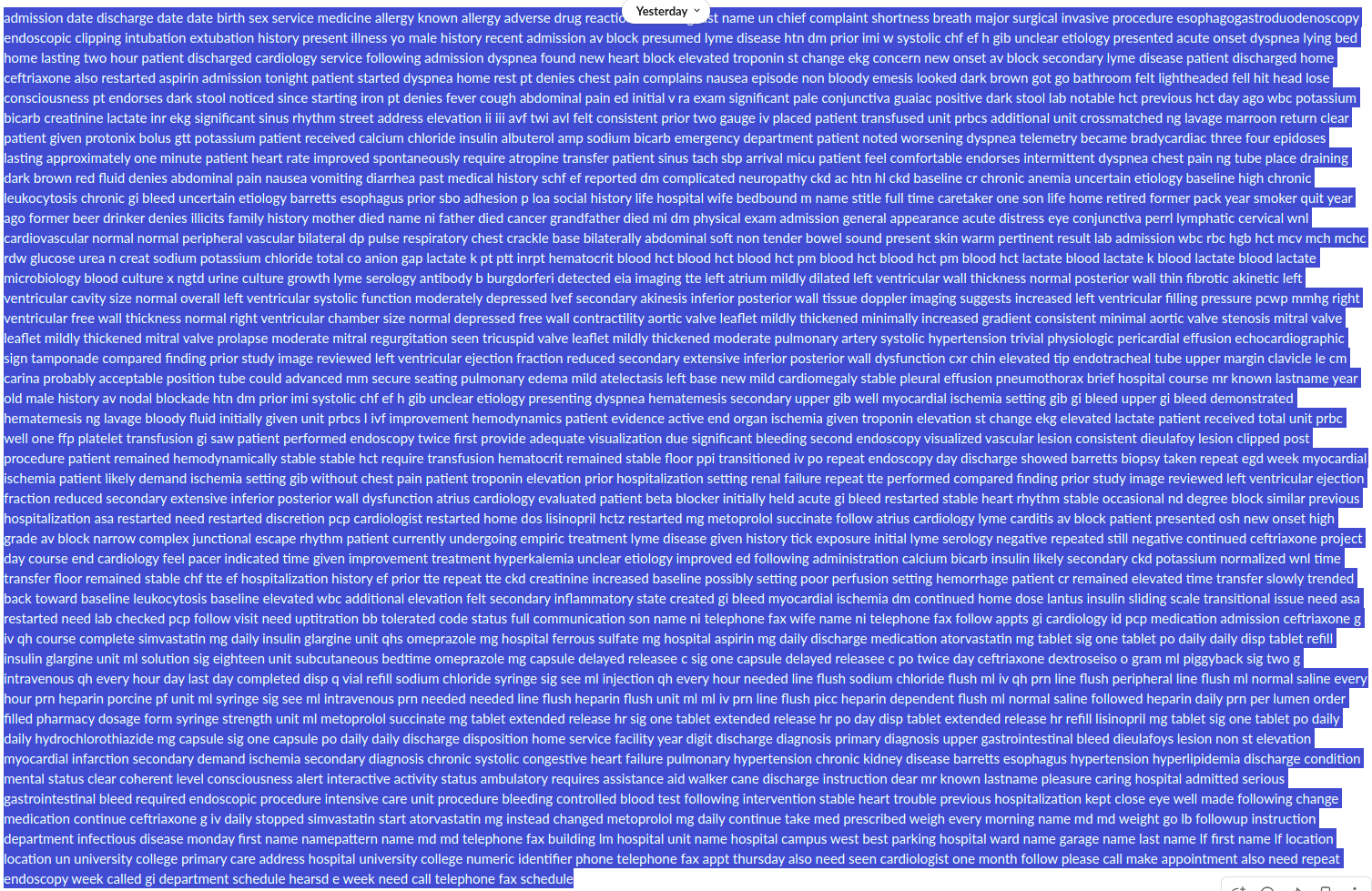}
    \caption{Processed Medical Notes}
    \label{fig:p3}
\end{figure}

The results indicate that the our application is effective in verifying medical notes, integrating real-time data for informed decision-making, and improving the user experience for healthcare workers. Most importantly, the application shows promise in its primary objective—reducing medical errors and enhancing patient safety.

\section{Limitation and Future directions}
In this project, we introduce an application aimed at reducing medical errors and enhancing patient safety, particularly in low-resource settings. We demonstrate the app's capability to verify medical notes and leverage openFDA data for informed decision-making.

However, our work has limitations. The application's effectiveness is currently tested on a limited dataset, questioning its generalizability. Additionally, the reliance on real-time data integration could pose challenges in regions with poor internet connectivity. As well, the FDA drug info might be too long for the LLM model due to the input token limit, which suggests that a summary may be required when the list of drugs is long. 

Future directions should focus on expanding the dataset for more robust testing and exploring offline capabilities to make the application more versatile. As well, there are models more specialized to medical texts, such as Google's Med-PaLM2, which is worth exploring once it becomes available. Additionally, reinforcement learning with human feedback (RLHF) can be added, in order to allow for model improvement based on doctor feedback, in the same fashion as ChatGPT. This work serves as a stepping stone for leveraging technology to reduce oversight in the judgement of healthcare professionals, this would especially improve healthcare outcomes in resource-poor settings.



\bibliographystyle{abbrv}
\bibliography{neurips}





\end{document}